\newif\ifisTR

\isTRtrue

\documentclass{article} % For LaTeX2e

\usepackage{fullpage}

\usepackage{microtype}
\usepackage{graphicx}
\usepackage{booktabs} % for professional tables
\usepackage{placeins}
\usepackage[linesnumbered,ruled,vlined]{algorithm2e}
\usepackage{algorithmic}
\usepackage{subcaption}
\usepackage{tcolorbox}
\usepackage{nicefrac}  
\usepackage{enumitem}
\usepackage{multirow}
\usepackage{colortbl}
\usepackage{xspace}
\usepackage{wrapfig}
\usepackage{amsmath}
\usepackage{amssymb}
\usepackage{mathtools}
\usepackage[export]{adjustbox}
\usepackage{amsthm}
\usepackage{xcolor}

\definecolor{CColor}{rgb}{0.01,0.31,0.59}
\definecolor{GGray}{rgb}{0.80,0.90,1}
\definecolor{Shady}{rgb}{0.9,0.9,0.9}
\definecolor{kaistblue}{RGB}{20,135,200}
\definecolor{kaistdarkblue}{RGB}{0,65,145}
\definecolor{urbanablue}{RGB}{19,41,75}
\definecolor{urbanaorange}{RGB}{232,74,39}
\definecolor{drp}{rgb}{0.53,0.15,0.34}
\usepackage[plainpages=false,pdfpagelabels,colorlinks=true,linkcolor=CColor,citecolor=CColor,urlcolor=CColor]{hyperref}

\usepackage{tikz}

\usepackage[capitalize,noabbrev]{cleveref}

\usepackage{overpic}

\theoremstyle{plain}
\newtheorem{theorem}{Theorem}[section]

\newtheorem{lemma}[theorem]{Lemma}

\theoremstyle{definition}

\theoremstyle{remark}

\definecolor{mygray}{gray}{0.85}
\definecolor{LightBlue}{cmyk}{0.06, 0.03, 0.01, 0.0}

\usepackage[textsize=tiny]{todonotes}

\usepackage[round,semicolon,sort]{natbib}

\renewcommand{\cite}[1]{\citep{#1}}

\date{}

\title{StructPrune: Structured Global Pruning asymptotics with $\mathcal{O}(\sqrt{N})$ GPU Memory}

\author{
 Xinyuan Song\textsuperscript{1},
 Guangji Bai\textsuperscript{1,2},
 Liang Zhao\textsuperscript{1,$\dagger$}
\\
 \textsuperscript{1}Emory University, USA\\
 \textsuperscript{2}Amazon, USA \\
 \textsuperscript{$\dagger$} Corresponding Author 
}

\begin{document}

\maketitle

\begin{abstract}
Pruning is critical for scaling large language models (LLMs). Global pruning achieves strong performance but requires $\mathcal{O}(N)$ memory, which is infeasible for billion-parameter models. Local pruning reduces GPU memory usage to that of a single layer by pruning layers independently, but it neglects inter-layer dependencies and often leads to suboptimal performance in high-sparsity regimes. Unlike unstructured pruning, structured pruning produces regular sparsity patterns that align well with GPU kernels and library optimizations, making it more hardware-efficient. However, structured pruning typically relies on global pruning, since structured patterns are more prone to severe performance degradation under local optimization. To jointly achieve structured pruning and the memory efficiency of local pruning, we propose a divide-and-conquer strategy that decomposes the global pruning problem into coordinated subproblems across different modules, each of which fits within limited GPU memory. Building on this idea, we design \textbf{STRUPRUNE}, an ADMM-based framework that integrates structured sparsity into the pruning process, combining the memory efficiency of local pruning with the hardware compatibility of structured methods. We derive a closed-form analytical solution for structured pruning masks that provides an explicit rule for layer-wise sparsity allocation, and further develop an energy-based asymptotic framework yielding a softmax-form allocation scheme that simplifies optimization while adapting to heterogeneous layer importance. Experiments demonstrate that STRUPRUNE matches the perplexity of global structured pruning while reducing memory cost from $\mathcal{O}(N)$ to $\mathcal{O}(\sqrt{N})$, enabling practical deployment at the billion-parameter scale.
\end{abstract}

\section{Introduction}
\label{sec:introduction}

Model pruning has proven effective in applications in computer vision and smaller language models~\cite{hoefler2021sparsity}. With the rapid development of large language models (LLMs), pruning has emerged as a key technique for reducing the computational and memory demands of billion-scale models~\cite{xu2023survey}. Classical global pruning approaches require loading the entire model into GPU memory, resulting in $\mathcal{O}(N)$ memory usage, scaling with the total number of parameters $N$~\cite{mallya2018packnet,singh2020woodfisher}. This requirement quickly becomes infeasible for modern LLMs containing tens to hundreds of billions of parameters.  

\begin{figure}[!ht]
 % 减小表格上方空间
    \centering
    \resizebox{0.8\columnwidth}{!}{
        \includegraphics[width=\linewidth]{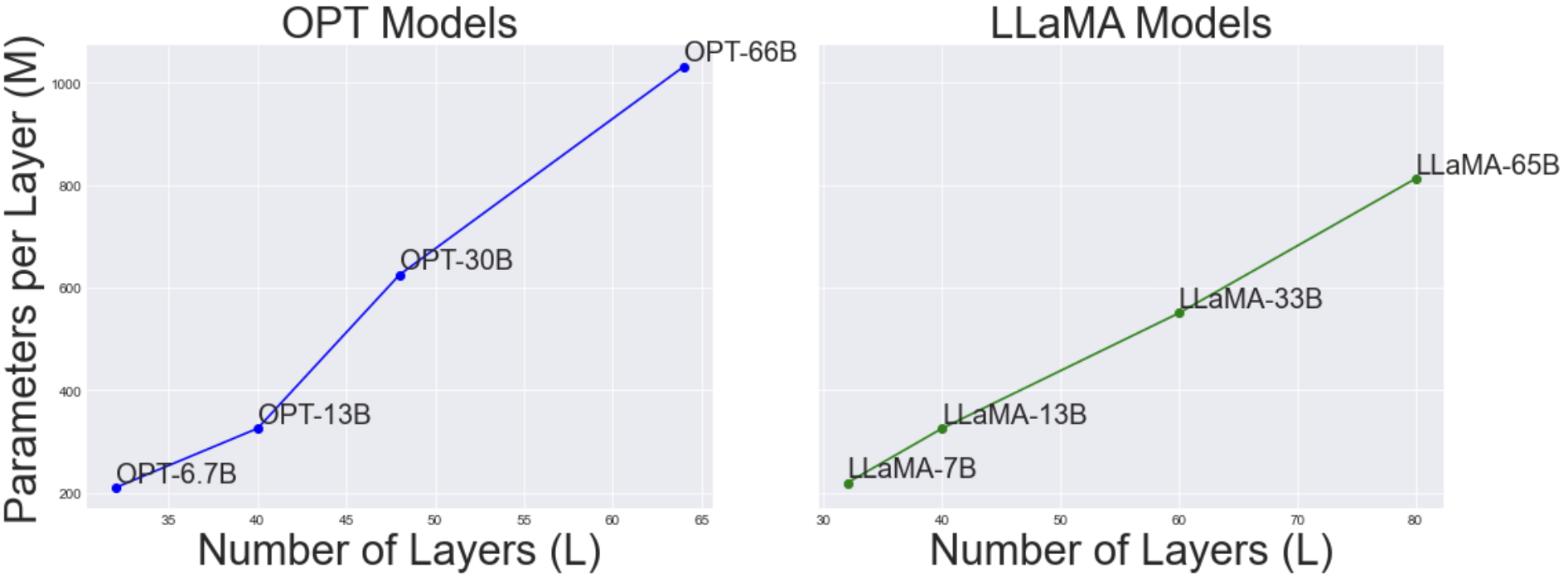}}

    \caption{Depth (number of layers) and width (parameters per layer) grow roughly proportionally as model size increases, for both OPT and LLaMA-2.}
    \label{fig:combined}
    
\end{figure}

To address the memory bottleneck, recent work has focused on \textbf{local pruning} methods~\cite{frantar2023massive,sun2023simple}, which prune each layer independently and then stitch the compressed layers together. This reduces the GPU footprint from the full model to that of a single layer, an advantage that becomes increasingly significant as LLMs scale up. As model size grows, both the number of layers and the parameters per layer increase roughly proportionally, with approximately $\mathcal{O}(\sqrt{N})$ layers and $\mathcal{O}(\sqrt{N})$ parameters per layer (see Table~\ref{tab:opt-model-size} and Figure~\ref{fig:combined}). For instance, SparseGPT~\cite{frantar2023massive} prunes LLMs with hundreds of billions of parameters while retaining competitive accuracy, and Wanda~\cite{sun2023simple} introduces a pruning criterion that accounts for both weight magnitude and input activation.  

However, local pruning inherently focuses on minimizing the error within each individual layer while ignoring inter-layer dependencies. As a result, it often yields suboptimal performance, particularly in high-sparsity regimes~\cite{singh2020woodfisher}. To address this limitation, \textbf{SparseLLM}~\cite{bai2024sparsellmglobalpruningpretrained} employs an Alternating Direction Method of Multipliers (ADMM)-based~\cite{boyd2011admm} algorithm to explicitly model and update inter-layer dependencies, thereby transforming the problem from isolated local optimization into coordinated global optimization. This design can yield substantial performance improvements in high-sparsity regimes.  

It is important to emphasize that these local approaches are primarily based on \textbf{unstructured pruning}, where sparsity patterns appear irregular at the weight level. Although this yields high compression ratios, the resulting irregular memory access severely limits acceleration on standard GPUs and TPUs, which are optimized for dense matrix operations~\cite{hoefler2021sparsity}. In practice, unstructured sparsity often requires custom kernels or specialized hardware support to achieve speedups. By contrast, \textbf{structured pruning} produces regular sparsity patterns by removing entire filters, channels, or blocks~\cite{he2017channel,ma2023llmpruner,sun2023simple,yuan2023gisp}, which map directly onto existing hardware primitives such as the Basic Linear Algebra Subprograms (BLAS) library~\cite{hoefler2021sparsity}.  

In this work, we propose \textbf{STRUPRUNE}, an ADMM-based divide-and-conquer framework that achieves global structured pruning while retaining the memory efficiency of local pruning by leveraging the ADMM method originally introduced in SparseLLM~\cite{bai2024sparsellmglobalpruningpretrained}. Specifically, we extend SparseLLM by replacing its unstructured pruning with structured pruning. This formulation reduces the memory requirement from $\mathcal{O}(N)$ to $\mathcal{O}(\sqrt{N})$ for structured pruning.  

Nevertheless, under structured pruning, \textbf{layer importance} becomes a decisive factor. Retaining critical layers, heads, and columns is far more meaningful than treating all components uniformly~\cite{yuan2023gisp,singh2020woodfisher}. The curse of depth phenomenon~\cite{sun2025cursedepthlargelanguage} shows that deeper layers often become redundant. Empirical evidence from LLM-Pruner~\cite{ma2023llmpruner} further demonstrates that layers contribute unequally to model performance. To address this, we first derive a closed-form analytical solution for the sparsity ratio of each layer, and then introduce an energy-based approximation that provides a theoretically reliable simplification. Both the analytical formulation and the asymptotic approximation are supported by rigorous mathematical analysis, ensuring that structured pruning can be applied consistently across layers. Extensive experiments show that our structured pruning framework maintains perplexity close to unpruned or benchmark baselines, while reducing GPU memory usage to that of a single layer compared to existing structured pruning methods.  

\begin{table}[!ht]
 % 减小表格上方空间
\centering
\caption{Parameter distribution and memory usage per layer for OPT models in FP16. \#Layers = Number of Layers; Params/L = Parameters per Layer; Mem/L = Memory per Layer.}
\renewcommand{\arraystretch}{0.8} % 行间距稍微缩小
\resizebox{0.7\columnwidth}{!}{
\begin{tabular}{lcccc}
\toprule
\midrule
Model & Tot. Params & \#Layers & Params/L (M) & Mem/L (MB) \\
\midrule
OPT-125M & 0.125B & 12 & 10.4   & 20.8   \\
OPT-350M & 0.350B & 24 & 14.6   & 29.2   \\
OPT-1.3B & 1.3B   & 24 & 54.2   & 108.4  \\
OPT-2.7B & 2.7B   & 32 & 84.4   & 168.8  \\
OPT-6.7B & 6.7B   & 32 & 209.4  & 418.8  \\
OPT-13B  & 13B    & 40 & 325.0  & 650.0  \\
OPT-30B  & 30B    & 48 & 625.0  & 1250.0 \\
OPT-66B  & 66B    & 64 & 1031.2 & 2062.4 \\
\midrule
\bottomrule
\end{tabular}}
\label{tab:opt-model-size}

\end{table}

\section{Related Work}
\label{sec:related_work}

Recent research has explored a wide range of structured pruning frameworks for LLMs. The Global Iterative Structured Pruning (GISP) framework~\cite{yuan2023gisp} decomposes pruning into iterative steps with progressively increasing sparsity, using normalized Taylor-based importance scores to balance attention and MLP modules, and further adapts the scoring function to task-specific objectives. Beyond iterative schemes, dynamic and architecture-aware approaches such as ToMoE~\cite{gao2025tomoeconvertingdenselarge} convert MLP layers into a differentiable Mixture-of-Experts to reduce active parameters without permanent deletion, while Tyr-the-Pruner~\cite{li2025tyrtheprunerunlockingaccurate50} constructs a supernet with diverse sparsity configurations, applying a coarse-to-fine search strategy to optimize sparsity allocation and achieve near-lossless performance at high pruning ratios. Block-wise and re-initialization-based methods further enrich the landscape: Thanos~\cite{ilin2025thanosblockwisepruningalgorithm} introduces an adaptive $N$:$M$ structured pruning algorithm that enables hardware-friendly block sparsity, and Pangu Light~\cite{chen2025pangulightweightreinitialization} combines weight re-initialization with pruning across width, depth, and attention heads to improve recovery after pruning. Finally, optimization-driven formulations such as SPAP~\cite{hu2025spapstructuredpruningalternating} cast pruning as a mixed-integer problem and employ penalty and alternating minimization strategies to achieve efficient compression with minimal accuracy degradation. These works collectively demonstrate the rapid progress of structured pruning in LLMs, spanning iterative, dynamic, block-wise, and optimization-based approaches.

\section{Global Structured Pruning}
\subsection{Optimization Objective}\label{one_layer_a_time_structured_pruning_method}

We adopt the ADMM-based formulation in SparseLLM but replace its unstructured pruning masks with structured pruning masks. Specifically, we define
\begin{equation}\label{eq:goal}
\begin{aligned}
    \min_{M_\ell,r_\ell} \; \mathcal{L}_{\ell}(\cdot; M \odot \widehat{W}) 
    &= \frac{1}{N} \sum_{i=1}^N \Big( 
        \alpha \big\| z_\ell^{\text{pre}} - (M_{\ell+1} \odot \widehat{W}_{\ell+1}) a_{\ell} \big\|_2^2  
        + \alpha \big\| W_\ell a_{\ell-1} - (M_\ell \odot \widehat{W}_{\ell}) a_{\ell-1}^{\text{pre}} \big\|_2^2 
    \Big)  \\
\text{s.t.} \quad  \sum_{\ell=1}^L \|M_{\ell}\|_1 &= \Big(\sum_{\ell=1}^L r_\ell\Big) n = r n L, \quad M_{\ell} \in \{0, 1\}^n.
\end{aligned}
\end{equation}
where $f(\cdot; M_\ell \odot \widehat{W})$ denotes the pruned network. Here, $M_\ell$ is the binary pruning mask applied at layer $\ell$, $\widehat{W}_\ell$ is the original weight matrix, $a_\ell$ and $a_{\ell-1}$ are the activations at layers $\ell$ and $\ell-1$, and $a_\ell^{\text{pre}}$ and $z_\ell^{\text{pre}}$ denote their corresponding pre-trained intermediate values (as in Equation~(4) of SparseLLM). The hyperparameter $\alpha$ balances reconstruction consistency across layers, $r_\ell$ is the sparsity ratio for layer $\ell$, $N$ is the number of samples in the calibration dataset, and $n$ is the number of structured units considered. Both $r_\ell$ and $M_\ell$ are optimization variables determined during structured pruning.

The derivation of our structured variant of SparseLLM~\cite{bai2024sparsellmglobalpruningpretrained}, including the decomposition of the optimization problem into subproblems and their closed-form or iterative solutions for both FFN and MHA modules, is presented in Algorithm~\ref{alg:sparse_llm} and Appendix~\ref{subproblem}. In particular, we detail how weight pruning, activation updates, and output updates are jointly optimized under structured constraints, and present the final algorithm with pseudocode. 

Since SparseLLM~\cite{bai2024sparsellmglobalpruningpretrained} focuses on updating and optimizing pruned matrices rather than investigating pruning strategies, we further derive a closed-form solution for structured pruning. Based on the objective in Equation~\eqref{eq:goal}, we now provide an analytical solution for the layer-wise sparsity and the induced binary mask.

\begin{lemma}[Layer-wise Sparsity and Mask Construction]\label{lemma}
For each layer $\ell$, solving the structured pruning objective gives the layer-wise sparsity ratio
\begin{equation}\label{rl}
r_\ell = \frac{1}{N}\sum_{j=1}^N \frac{c_j b_j + d_j z^{\rm pre}_{\ell-1,j}}{c_j^2+d_j^2},
\end{equation}
where the intermediate variables are defined as
\begin{equation}
b_j = [W_\ell a_{\ell-1}]_j,\quad
c_j = [\widehat W_\ell a_{\ell-1}^{\mathrm{pre}}]_j,\quad
d_j = [\widehat W_{\ell+1} a_\ell]_j.
\end{equation}
For binary masks $M_{\ell,j} \in \{0,1\}$, we define an importance score for each unit:
\begin{equation}
s_j = \frac{c_j b_j + d_j z^{\rm pre}_{\ell-1,j}}{c_j^2+d_j^2},\quad 
M_{\ell,j} = \begin{cases}
1, & \text{if } s_j > \tau,\\
0, & \text{otherwise.}
\end{cases}
\end{equation}
\end{lemma}

The threshold $\tau$ is chosen so that the number of retained units matches the target sparsity ratio. The detailed proof is provided in Section~\ref{111222}. Thus, both the binary mask $M_{\ell}$ and the sparsity ratio $r_\ell$ are jointly determined by ranking the scores $\{s_j\}$, yielding a consistent rule for structured pruning across layers.

Algorithm~\ref{alg:GSPO} in Section~\ref{111222} presents the complete procedure for solving the optimization problem in Equation~\eqref{eq:goal}, combining the analytical sparsity ratio, score-based mask construction, and iterative updates into a unified workflow.

\subsection{Energy-Based Asymptotic for Continuous Approximation}

In many structured sparsification and pruning schemes, the optimization over binary gating variables \( M_{\ell,j} \in \{0,1\} \) at each layer \( \ell \) can be computationally intractable or statistically unnecessary. Rather than optimizing over all binary configurations, we focus on the layer-wise retention rate \( r_\ell = \frac{1}{n} \sum_{j=1}^n M_{\ell,j} \in [0,1] \), which specifies the fraction of active units at that layer and serves as a continuous approximation of the binary mask. This motivates a relaxation of Lemma~\ref{lemma} into a continuous mean-field form.

Building on the analytical formulation above, we propose an energy-based asymptotic framework. Theoretical analysis shows that this asymptotic approximation is consistent with, and closely approximates, the exact solution presented earlier.

From Equation~\eqref{eq:goal}, we aim to minimize the total energy
\begin{equation}\label{qwe}
    E_{\mathrm{tot}} = \sum_{\ell=1}^L E_\ell(r_\ell), \quad \text{s.t.}\sum_{\ell=1}^L r_\ell = rL,
\end{equation}
where each $E_\ell(r_\ell)$ is a differentiable energy function representing the right-hand side of Equation~\eqref{eq:goal}, and $r_\ell \in (0,1]$ denotes the fraction of units retained in layer $\ell$. Introducing a Lagrange multiplier $\lambda$ to enforce the equality constraint yields
\begin{equation}\label{qwer}
    \mathcal{L} = \sum_{\ell=1}^L E_\ell(r_\ell) + \lambda \left( \sum_{\ell=1}^L r_\ell - rL \right).
\end{equation}

\begin{lemma}[Asymptotic Layer-wise Sparsity Allocation]\label{thm:sparsity}
Building on Lemma~\ref{lemma}, the structured pruning objective can be reformulated into the form of Equations~\eqref{qwe} and~\eqref{qwer}. Under this formulation, the optimal layer-wise sparsity ratio in Equation~\eqref{rl} has the asymptotic closed-form solution
\begin{equation}
    r_\ell^* = rL \cdot \mathrm{softmax}\!\left(-\frac{I_\ell}{T}\right),
\end{equation}
where $I_\ell$ denotes the importance score of layer $\ell$ defined in Lemma~\ref{lemma}, and $T$ is a temperature parameter controlling the sharpness of the allocation.
\end{lemma}

The proof of Lemma~\ref{thm:sparsity} and the mathematical background are provided in Section~\ref{energy}. The benefit of this asymptotic characterization is that it yields a closed-form allocation rule, which greatly simplifies optimization by avoiding iterative constrained solvers while retaining adaptivity to heterogeneous layer importance. Furthermore, Algorithms~\ref{alg:importance-pruning} and~\ref{alg:importance-pruning-distribute} in Section~\ref{energy} are designed based on this asymptotic formulation: Algorithm~\ref{alg:importance-pruning} implements a baseline variant with uniform layer-wise sparsity allocation, while Algorithm~\ref{alg:importance-pruning-distribute} introduces asymmetric importance estimation and depth-aware sparsity allocation. Details are provided in Section~\ref{energy}.

\subsection{Algorithmic Details}
\label{appendix:algorithms}
In this section, we provide detailed descriptions of the pruning algorithms based on our energy-based framework. Algorithm~\ref{alg:importance-pruning} presents a baseline variant with uniform layer-wise sparsity allocation, in which pruning is guided solely by global importance scores. In contrast, Algorithm~\ref{alg:importance-pruning-distribute} extends this design by separating attention and MLP modules and introducing a depth-aware decay factor. Together, these two variants illustrate the progression from a simple, globally uniform strategy to a more flexible, fine-grained scheme that better captures heterogeneous sensitivity across components. 

 \begin{algorithm*}[!ht]
 \footnotesize
\caption{Importance-Guided Global Structured Pruning with Temperature Optimization}
\label{alg:importance-pruning}
\begin{algorithmic}[1]
\STATE \textbf{Input:} Calibration dataset $\mathcal{D} = \{(x_i, y_i)\}_{i=1}^N$, pretrained weights $\widehat{W}$, global sparsity target $\bar{r}$, number of layers $L$
\STATE \textbf{Output:} Structured sparse model with optimized pruning masks and SparseLLM scheduling
\FOR{temperature $T$ in predefined search grid $\mathcal{T}$}
    \STATE \textbf{Step 1: Compute Layerwise Importance}
    \FOR{$\ell = 1$ to $L$}
        \STATE Estimate importance score $I_\ell$ (e.g., WANDA, gradient-based, etc.)
    \ENDFOR
    \STATE \textbf{Step 2: Softmax-based Sparsity Allocation with Post-correction}
    \STATE Normalize importance scores with negative softmax:
    \begin{equation}
    w_\ell = \frac{\exp(-I_\ell / T)}{\sum_{j=1}^L \exp(-I_j / T)}
    \end{equation}
    \STATE Allocate raw sparsity:
    \begin{equation}
    r_\ell = \bar{r} \cdot L \cdot w_\ell
    \end{equation}
    \STATE Clip: $r_\ell \leftarrow \min(\max(r_\ell, 0), 0.95)$
    \IF{no $r_\ell$ was clipped to $0.95$}
        \STATE Compute current mean: $\mu = \frac{1}{L} \sum_{\ell=1}^L r_\ell$
        \STATE Rescale: $r_\ell \leftarrow r_\ell \cdot \frac{\bar{r}}{\mu}$
        \STATE Final clip: $r_\ell \leftarrow \min(\max(r_\ell, 0), 0.95)$
    \ENDIF
    \STATE \textbf{Step 3: Apply Wanda Structured Pruning}
    \FOR{$\ell = 1$ to $L$}
        \STATE Compute Wanda score:
        \begin{equation}
        s_i^{(\ell)} = \frac{1}{|\mathcal{D}|} \sum_{x \in \mathcal{D}} \left\| W_i^{(\ell)} \cdot x_i \right\|_1
        \end{equation}
        \STATE Generate binary mask $M_\ell$ with sparsity $r_\ell$ via score thresholding
        \STATE Apply pruning: $\widetilde{W}_\ell = M_\ell \odot \widehat{W}_\ell$
    \ENDFOR
    \STATE \textbf{Step 4: SparseLLM Optimization}
    \STATE Feed $\{\widetilde{W}_\ell\}_{\ell=1}^L$ to SparseLLM for dependency scheduling
    \STATE \textbf{Step 5: Evaluate Final Loss}
    \begin{equation}
    \mathcal{L}(T) = \sum_{\ell=1}^L \mathcal{L}_\ell(\cdot; r_\ell, T)
    \end{equation}
\ENDFOR
\STATE \textbf{Return:} Best pruned model and $T^\star$ minimizing $\mathcal{L}(T)$
\end{algorithmic}
 \end{algorithm*}

 \begin{algorithm*}[!ht]
 \footnotesize
\caption{Importance-Guided Global Structured Pruning with Temperature Optimization}
\label{alg:importance-pruning-distribute}
\begin{algorithmic}[1]
\STATE \textbf{Input:} Calibration dataset $\mathcal{D} = \{(x_i, y_i)\}_{i=1}^N$, pretrained weights $\widehat{W}$, global sparsity target $\bar{r}$, number of layers $L$, attention/MLP importance ratio $\rho$, depth decay factor $\gamma$
\STATE \textbf{Output:} Structured sparse model with optimized pruning masks and SparseLLM scheduling
\FOR{temperature $T$ in predefined search grid $\mathcal{T}$}
    \STATE \textbf{Step 1: Estimate Layer-wise Importance}
    \FOR{$\ell = 1$ to $L$}
        \STATE Compute depth decay weight: $d_\ell = \gamma^\ell$
        \IF{Layer $\ell$ is MHAttention}
            \STATE Compute attention importance:
            \begin{equation}
            I_\ell^{\text{attn}} = d_\ell \cdot \rho \cdot \sum_{h \in \text{AttnHeads}(\mathcal{L}_\ell)} \left( \|W_h\|_1 + 0.1 \cdot \text{Var}(W_h) + 0.01 \cdot \|W_h\|_2 \right)
            \end{equation}
        \ELSIF{Layer $\ell$ is MLP}
            \STATE Compute MLP importance:
            \begin{equation}
            I_\ell^{\text{mlp}} = d_\ell \cdot \sum_{m \in \{fc1, fc2\}} \left( \|W_m\|_1 + 0.05 \cdot \text{Var}(W_m) \right)
            \end{equation}
        \ENDIF
    \ENDFOR

    \STATE \textbf{Step 2: Compute Layer-wise Sparsity via Temperature-Guided Inverse Weighting}
    \STATE Normalize attention and MLP importance separately:
    \begin{equation}
    w_\ell^{\text{attn}} = \frac{\exp(-I_\ell^{\text{attn}} / T)}{\sum_j \exp(-I_j^{\text{attn}} / T)}, \quad
    w_\ell^{\text{mlp}} = \frac{\exp(-I_\ell^{\text{mlp}} / T)}{\sum_j \exp(-I_j^{\text{mlp}} / T)}
    \end{equation}
    \STATE Compute inverse-normalized sparsity:
    \begin{equation}
    r_\ell^{\text{attn}} = \text{clip}\left(\bar{r} \cdot L \cdot \frac{1 - w_\ell^{\text{attn}}}{\sum_j (1 - w_j^{\text{attn}})}, 0, 0.95\right)
    \end{equation}
    \begin{equation}
    r_\ell^{\text{mlp}} = \text{clip}\left(\bar{r} \cdot L \cdot \frac{1 - w_\ell^{\text{mlp}}}{\sum_j (1 - w_j^{\text{mlp}})}, 0, 0.95\right)
    \end{equation}

    \STATE \textbf{Step 3: Apply Wanda Structured Pruning}
    \FOR{$\ell = 1$ to $L$}
        \STATE Compute Wanda score:
        \begin{equation}
        s_i^{(\ell)} = \frac{1}{|\mathcal{D}|} \sum_{x \in \mathcal{D}} \left\| W_i^{(\ell)} \cdot x_i \right\|_1
        \end{equation}
        \STATE Generate binary mask $M_\ell$ with sparsity $r_\ell^{\text{attn}}$ or $r_\ell^{\text{mlp}}$ via score thresholding
        \STATE Apply pruning: $\widetilde{W}_\ell = M_\ell \odot \widehat{W}_\ell$
    \ENDFOR

    \STATE \textbf{Step 4: SparseLLM Optimization}
    \STATE Feed $\{\widetilde{W}_\ell\}_{\ell=1}^L$ to SparseLLM for dependency scheduling

    \STATE \textbf{Step 5: Evaluate Final Loss}
    \begin{equation}
    \mathcal{L}(T) = \sum_{\ell=1}^L \mathcal{L}_\ell(\cdot; r_\ell, T)
    \end{equation}
\ENDFOR
\STATE \textbf{Return:} Best pruned model and $T^\star$ minimizing $\mathcal{L}(T)$
\end{algorithmic}
 \end{algorithm*}

Algorithm~\ref{alg:importance-pruning-distribute} differs from Algorithm~\ref{alg:importance-pruning} in two key aspects: importance estimation and sparsity allocation. First, instead of computing a single unified importance score for each layer, it explicitly separates attention and MLP modules, assigning each an independent importance measure. For attention, importance is derived from the $\ell_1$ norm, variance, and $\ell_2$ norm of each attention head’s weight matrix. These values are further scaled by a global attention-to-MLP importance ratio $\rho$, which reflects the higher pruning sensitivity of attention layers. In contrast, MLP importance is calculated using the $\ell_1$ norm and variance of fully connected layers $\text{FC\_1}$ and $\text{FC\_2}$, without additional amplification. Second, to account for depth-related relevance decay, a geometric factor $\gamma^\ell$ is applied to both attention and MLP scores, thereby down-weighting the importance of deeper layers. This design encourages more aggressive pruning of later-stage layers, under the assumption that deeper layers tend to exhibit greater redundancy. Together, these changes allow Algorithm~\ref{alg:importance-pruning-distribute} (implemented in \texttt{opt\_main\_dist.py}) to provide finer-grained, asymmetric sparsity control between functional modules and across depth, in contrast to the more uniform treatment of Algorithm~\ref{alg:importance-pruning}.

\subsection{Layer Importance \texorpdfstring{$I_\ell$}{} Estimation}

Several criteria have been proposed to estimate the importance of layers in large models. Gradient-based sensitivity methods such as SNIP~\cite{lee2018snip} estimate pruning impact via first-order loss perturbations, while Hessian-based approaches (e.g., OBD/OBS~\cite{singh2020woodfisher}) exploit second-order curvature information. Feature reconstruction error~\cite{he2017channel} has also been used to measure the discrepancy between pre- and post-pruning hidden representations. Although these methods provide theoretical insight, they are often computationally expensive or numerically unstable at scale. In contrast, we adopt the simplest and most practical option, \textbf{Wanda structured pruning}~\cite{wanda2023}, since it requires minimal extra computation while achieving effective structured sparsity guided by input norms. Specifically, Wanda defines the importance score as
\begin{equation}\label{wanda}
I(W) = \|W\|_2 \odot \left( \mathbf{1} \cdot \| \mathbf{X}_{\text{in}} \|_2^\top \right),
\end{equation}
where $\mathbf{X}_{\text{in}}$ denotes the input activations, $\|\cdot\|_2$ is the $\ell_2$ norm, $\mathbf{1}$ is an all-ones vector, and $\odot$ denotes the Hadamard product. This lightweight strategy has proven efficient and well suited for large language models.

\subsection{Update of Remaining Weights}

The remaining weights $W_\ell$ after pruning can be updated using several approaches. Several matrix recovery techniques have been explored, including column-row decomposition ($M=CUR$)~\cite{drineas2006cur}, pivoted LU decomposition~\cite{golub2013matrix}, interpolation-based methods~\cite{cheng2005compression}, and adaptive cross approximation~\cite{bebendorf2003adaptive}. However, structured pruning often produces non-invertible matrices, which limits the applicability of these closed-form solutions. In our method, we adopt LoRA~\cite{hu2021lora}, a gradient descent fine-tuning strategy, to update the remaining weights effectively.

\section{Experiments}

\subsection{Models and Datasets}

We evaluate our method on OPT-125M~\cite{opt2024}. Experiments are conducted on a single NVIDIA A24 GPU with 24~GB of memory. Following~\cite{frantar2023massive}, we select 128 random 2048-token segments from the first shard of the C4 dataset as calibration data. This ensures zero-shot pruning without task-specific fine-tuning. 

The OPT model series—including OPT-125M, OPT-350M, and larger variants—adheres to a consistent Transformer architecture in which the parameter ratio between the multi-head attention (MHA) module and the feed-forward network (FFN) remains approximately \(1{:}2\) across all model scales. This design arises because the FFN expands the hidden size \(d\) to \(4d\), resulting in two linear projections of size \(d \times 4d\) and \(4d \times d\), contributing \(8d^2\) parameters per layer. In contrast, the MHA module uses four projection matrices (query, key, value, and output), each of size \(d \times d\), totaling \(4d^2\) parameters. Hence, the FFN-to-MHA parameter ratio is consistently \(\frac{8d^2}{4d^2} = 2{:}1\).

As shown in Table~\ref{tab:opt_ratios}, this ratio is invariant to model size, ensuring a balanced allocation between attention and feed-forward components. Empirical inspection of OPT models confirms this architectural consistency, with both MHA and FFN parameter counts scaling quadratically with \(d\), while preserving their relative proportion.
\begin{table*}[!ht]
\centering
\caption{Parameter Distribution Between FFN and MHA in OPT Models}
\label{tab:opt_ratios}
\resizebox{0.8\columnwidth}{!}{
\begin{tabular}{@{}lcccc@{}}
\toprule
\midrule
\textbf{Model} & \textbf{Hidden Size \(d\)} & \textbf{FFN Params (\(8d^2\))} & \textbf{MHA Params (\(4d^2\))} & \textbf{FFN:MHA Ratio} \\
\midrule
OPT-125M  & 768   & 4,718,592   & 2,359,296   & 2.00 \\
OPT-350M  & 1,024 & 8,388,608   & 4,194,304   & 2.00 \\
OPT-1.3B  & 2,048 & 33,554,432  & 16,777,216  & 2.00 \\
OPT-2.7B  & 2,560 & 52,428,800  & 26,214,400  & 2.00 \\
OPT-6.7B  & 4,096 & 134,217,728 & 67,108,864  & 2.00 \\
OPT-13B   & 5,120 & 209,715,200 & 104,857,600 & 2.00 \\
OPT-30B   & 7,168 & 411,041,792 & 205,520,896 & 2.00 \\
OPT-66B   & 9,216 & 679,477,248 & 339,738,624 & 2.00 \\
\midrule
\bottomrule
\end{tabular}}
\end{table*}

For evaluation, we primarily report perplexity (PPL) on the WikiText-2 and C4 validation subsets, since it is widely recognized as a robust metric for compression methods~\cite{dettmers2023case}. As baselines, we include sliceGPT~\cite{ashkboos2024slicegpt} and FASP~\cite{hu2025fasp}, representing state-of-the-art structured pruning methods under comparable settings. We evaluate three variants of our proposed approach: (1) a closed-form base model that directly computes the layer-wise sparsity ratio (Algorithm~\ref{alg:GSPO}); (2) an asymptotic model with sparsity planning that adjusts allocation across layers (Algorithm~\ref{alg:importance-pruning}); and (3) an asymptotic model without explicit sparsity planning that applies the theoretical softmax-based allocation in simplified form (Algorithm~\ref{alg:importance-pruning-distribute}). To ensure consistent importance weighting across layers, the temperature \(T\) is scaled to the magnitude of the importance scores. 

\subsection{Main Results}

We report perplexity results for OPT-125M on WikiText-2 and C4 under different pruning variants, focusing on sparsity levels of 0.1, 0.2, and 0.3. The results are summarized in Table~\ref{tab:wanda_MLP_structured_dist_125}.

\begin{table}[!ht]
\centering
\caption{Perplexity of OPT-125M on Wikitext2 and C4 using C4 as calibration dataset}
\label{tab:wanda_MLP_structured_dist_125}
\resizebox{0.7\columnwidth}{!}{
\begin{tabular}{lccc|ccc}
\toprule
\midrule
\textbf{Method} & \multicolumn{3}{c|}{\textbf{Wikitext2 (PPL)}} & \multicolumn{3}{c}{\textbf{C4 (PPL)}} \\
 & 0.10 & 0.20 & 0.30 & 0.10 & 0.20 & 0.30 \\
\midrule
sliceGPT~\cite{ashkboos2024slicegpt} & 29 & 34 & 45 & 27 & 34 & 40 \\
FASP~\cite{hu2025fasp} & \textbf{28} & \textbf{30} & \textbf{34} & \textbf{26} & \textbf{28} & \textbf{37} \\
\midrule
Close form (Algorithm~\ref{alg:GSPO}) & 31 & 51 & 83 & 30 & 45 & 87 \\
Asymptotics (Algorithm~\ref{alg:importance-pruning})  & 30 & 46 & 73 & 28 & 39 & 57 \\
Asymptotics (Algorithm~\ref{alg:importance-pruning-distribute})  & 29 & 59 & 93 & 27 & 55 & 98 \\
\midrule
\bottomrule
\end{tabular}}

\end{table}

The results in Tables~\ref{tab:wanda_MLP_structured_dist_125} demonstrate that our pruning framework achieves consistently lower perplexity than the benchmark baselines under moderate sparsity levels. On OPT-125M, our variants preserve performance close to the unpruned model, particularly at low to medium sparsity. We further compare GPU memory consumption across different pruning methods on OPT models of varying sizes. Table~\ref{tab:memory_usage} reports the memory footprint during pruning for OPT-125M and OPT-1.3B, with our method listed first, followed by the structured pruning baselines sliceGPT and FASP.

\begin{table}[!ht]
\centering
\caption{GPU memory usage (GB) of the model from different pruning methods on OPT models.}
\label{tab:memory_usage}
\resizebox{0.7\columnwidth}{!}{
\begin{tabular}{lccc}
\toprule
\midrule
\textbf{Method} & \textbf{OPT-125M} & \textbf{OPT-350M} & \textbf{OPT-1.3B} \\
\midrule
FASP~\cite{hu2025fasp} & 0.24 & 0.69 & 2.51 \\
sliceGPT~\cite{ashkboos2024slicegpt} & 0.25 & 0.71 & 2.42 \\
\textbf{Ours} & \textbf{0.02} & \textbf{0.02} & \textbf{0.10} \\
\midrule
\bottomrule
\end{tabular}}

\end{table}
As shown in Table~\ref{tab:memory_usage}, our method reduces the GPU memory footprint by an order of magnitude compared to FASP and SliceGPT across all tested OPT scales. For instance, on OPT-1.3B, our framework requires only 0.10~GB, whereas the baselines exceed 2~GB. This substantial reduction highlights the memory advantage of our approach, enabling structured pruning to be applied efficiently even to billion-parameter LLMs under limited hardware budgets.

\subsection{Other Structured Pruning Methods}

In addition to our analytical formulations, we also evaluate several existing structured pruning methods applied within SparseLLM on OPT-125M, with results summarized in Table~\ref{tab:pruning_compare} at sparsity levels of 0.1, 0.2, and 0.3. The comparison covers four representative approaches: SNIP~\cite{lee2018snip}, L0 regularization~\cite{louizos2018l0}, Magnitude-based pruning~\cite{han2015learning}, and remaining weight correction. Benchmark perplexity values on WikiText-2 and C4 are also reported for reference and align well with our method, confirming the reliability of the evaluation setup. These results show that standard structured pruning methods, when applied without global optimization, tend to degrade rapidly as sparsity increases. In contrast, our approach degrades much more slowly, demonstrating greater robustness under high sparsity.
\begin{table}[!ht]
\centering
\caption{Perplexity of OPT-125M on Wikitext2 and C4 under different structured pruning methods at sparsity 0.1 to 0.3}
\label{tab:pruning_compare}
\resizebox{0.7\columnwidth}{!}{
\begin{tabular}{lccc|ccc}
\toprule
\midrule
\multirow{2}{*}{\textbf{Method}} & \multicolumn{3}{c|}{\textbf{Wikitext2 (PPL)}} & \multicolumn{3}{c}{\textbf{C4 (PPL)}} \\
 & 0.10 & 0.20 & 0.30 & 0.10 & 0.20 & 0.30 \\
\midrule
SNIP & 46 & 126 & 439 & 39 & 88 & 264 \\
SNIP (w/ correction) & 42 & 112 & 386 & 36 & 80 & 231 \\
\midrule
L0 & 101 & 315 & 677 & 94 & 297 & 634 \\
L0 (w/ correction) & 93 & 299 & 633 & 88 & 281 & 605 \\
\midrule
Magnitude & 49 & 133 & 2084 & 45 & 101 & 2197 \\
Magnitude (w/ correction) & 44 & 125 & 1997 & 41 & 93 & 2007 \\
\midrule
Close form (Algorithm~\ref{alg:GSPO}) & 31 & 51 & 83 & 30 & 45 & 87 \\
Asymptotics (Algorithm~\ref{alg:importance-pruning})  & 30 & \textbf{46} & \textbf{73} & 28 & \textbf{39} & \textbf{57} \\
Asymptotics (Algorithm~\ref{alg:importance-pruning-distribute})  & \textbf{29} & 59 & 93 & \textbf{27} & 55 & 98\\
\midrule
\bottomrule
\end{tabular}}

\end{table}

\section{Conclusion}
\label{sec:conclusion}

We introduced \textbf{STRUPRUNE}, a structured pruning framework that combines the memory efficiency of local pruning with global coordination via ADMM-based optimization. By reducing memory usage from $\mathcal{O}(N)$ to $\mathcal{O}(\sqrt{N})$, STRUPRUNE enables pruning of billion-parameter LLMs within practical GPU limits, while its structured masks provide hardware-friendly acceleration. A key contribution is the derivation of a closed-form analytical solution for layer-wise sparsity, along with an energy-based asymptotic approximation that provides a theoretically sound, computationally efficient alternative. Experiments on OPT models demonstrate that STRUPRUNE maintains perplexity close to global structured pruning while substantially lowering memory consumption, demonstrating its practicality and effectiveness for large-scale deployment.

\bibliographystyle{plainnat}
\bibliography{reference}

\clearpage
\appendix

\section{Derive subproblems \& solutions}\label{subproblem} 

In this appendix, we present the complete derivation and solution steps for our structured variant of SparseLLM. Building on the ADMM-based global pruning method proposed in~\cite{bai2024beyond}, we extend the formulation from unstructured pruning to structured pruning applied to both feed-forward (FFN) and multi-head attention (MHA) modules. The ADMM framework treats pruning as a constrained optimization problem and introduces auxiliary variables with alternating direction updates, thereby enabling global coordination of sparsity patterns across layers. Within this framework, we decompose the global optimization into subproblems for weight pruning, activation updates, and output reconstruction, each solvable by least-squares or gradient-based solvers. We further incorporate the reverse reconstruction of weight matrices and an outer optimization cycle that integrates these subproblems into a unified procedure. The full structured pruning workflow for OPT-style decoder layers is summarized in Algorithm~\ref{alg:sparse_llm}.

\subsection{Computation on FFN Layers}\label{FFN}
As a concrete case, we first detail how the ADMM-based pruning procedure is applied to the feed-forward network (FFN) layers within each decoder block of a pre-trained LLM. Specifically, for each FFN module, we minimize the following objective:
\begin{equation}
\begin{aligned}
    \alpha \|z_\ell^{\text{pre}} - (M_{\ell+1} \odot \widehat{W}_{\ell+1}) a_{\ell} \|_2^2 
    + \beta \| a_\ell - \phi_\epsilon (z_\ell) \|_2^2 
    + \alpha \| z_\ell - (M_\ell \odot \widehat{W}_{\ell}) a_{\ell-1}^{\text{pre}} \|_2^2.
    \end{aligned}
\end{equation}

The optimization procedure consists of the following steps:
\begin{itemize}[left = 0em]
    \item \textbf{Pruning weight.}  
    Optimize $M_\ell$ and $\widehat{W}_\ell$ with respect to the loss 
    \begin{equation}
        \| z_\ell - (M_\ell \odot \widehat{W}_{\ell}) a_{\ell-1} \|_2^2.
    \end{equation}  
    By decomposing $z_\ell = W_\ell a_{\ell-1}$, where $W_\ell = z_\ell a_{\ell-1}^\dagger$, the objective becomes  
    \begin{equation}
        \| W_\ell a_{\ell-1} - (M_\ell \odot \widehat{W}_{\ell}) a_{\ell-1} \|_2^2,
    \end{equation}  
    which can be solved using the method described in Section~\ref{one_layer_a_time_structured_pruning_method}.

    \item \textbf{Updating activation.}  
    The minimization for $a_\ell$ reduces to a least-squares problem:
    \begin{equation}
        a_{\ell} = (\alpha W_{\ell+1}^T W_{\ell+1} + \beta I)^{-1} 
        (\alpha W_{\ell+1}^T z_{\ell+1}^{\text{pre}} + \beta \cdot \text{ReLU}(z_\ell)).
    \end{equation}

    \item \textbf{Updating output.}  
    The update for $z_\ell$ requires minimizing
    \begin{equation}
        \beta \| a_\ell - \text{ReLU}(z_\ell) \|_2^2 
        + \alpha \| z_\ell - (M_\ell \odot \widehat{W}_{\ell}) a_{\ell-1}^{\text{pre}} \|_2^2.
    \end{equation}  
    The closed-form solution can be written as:
    \begin{equation}
        z_\ell^{(1)} = (M_\ell \odot \widehat{W}_{\ell}) a_{\ell-1}^{\text{pre}}, 
        \quad 
        z_\ell^{(2)} = (\alpha + \beta)^{-1} \cdot (\beta a_\ell + \alpha z_\ell^{(1)}).
    \end{equation}

    \item \textbf{Reversely recovering the weight matrix.}  
    Recover the weight matrices from activations using:
    \begin{equation}
        \mathbf{W}_\ell = \mathbf{z}_\ell a_{\ell-1}^\dagger, 
        \quad 
        \mathbf{W}_{\ell+1} = \mathbf{z}_{\ell+1} a_\ell^\dagger.
    \end{equation}  
    Since the pseudo-inverse may be difficult to compute in practice, we alternatively employ an SGD-based solver to optimize for $\mathbf{W}_\ell$.

    \item \textbf{Outer optimization loop.}  
    The optimized $\mathbf{W}_\ell$ and $\mathbf{W}_{\ell+1}$ are then fed back into the global cycle. By iteratively repeating this process, we obtain pruned weight matrices that balance sparsity and reconstruction fidelity.
\end{itemize}

In summary, the pruning process alternates between weight optimization, activation updates, and output reconstruction, while periodically recovering and refining the weight matrices. This iterative scheme ensures stable convergence and effective global structured pruning for large-scale LLMs.

\subsection{Computation on Multi-Head Attention layers}

Consider the $\ell$-th decoder layer, where the multi-head attention mechanism is given by (omitting batch/layer normalization and residual connections for clarity):

\begin{itemize}[left = 0em]
    \item Compute $Q_\ell, K_\ell, V_\ell$:
    \begin{equation}
        Q_\ell = W_\ell^Q  a^{pre}_{\ell-1}, \quad
        K_\ell = W_\ell^K  a^{pre}_{\ell-1}, \quad
        V_\ell = W_\ell^V  a^{pre}_{\ell-1}.
    \end{equation}
    \item Each head $i$ is computed as
    \begin{equation}
        z_{\ell}= \tfrac{Q_\ell (K_\ell)^\top}{\sqrt{d_{\mathrm{head}}}},
    \end{equation}
    \begin{equation}
        \mathbf{a}_\ell = \phi_\ell\Bigl(\tfrac{Q_\ell (K_\ell)^\top}{\sqrt{d_{\mathrm{head}}}}\Bigr),
    \end{equation}
    \begin{equation}
        \text{head}_i = \phi_\ell\Bigl(\tfrac{Q_\ell^i (K_\ell^i)^\top}{\sqrt{d_{\mathrm{head}}}}\Bigr)V_\ell^i,
    \end{equation}
    where $Q_\ell^i, K_\ell^i, V_\ell^i$ are slices of $Q_\ell, K_\ell, V_\ell$ in the column dimension. $\phi_l$ is the softmax activation function.
    \item Concatenate all heads and multiply by $W_\ell^O$:
    \begin{equation}
        a_\ell^{\mathrm{attn}} = \mathrm{Concat}(\mathrm{head}_1, \ldots, \mathrm{head}_h)
    \end{equation}
    we also can write
    \begin{equation}
        z_{\ell+1}^{pre} = \mathrm{Concat}(\mathrm{head}_1, \ldots, \mathrm{head}_h) W_\ell^O.
    \end{equation}
\end{itemize}

We aim to prune $\{W_\ell^Q,W_\ell^K,W_\ell^V,W_\ell^O\}$ with minimal distortion to the pre-trained outputs $z_\ell^{\text{pre}}$, in a manner analogous to the FFN pruning.

\subsection{Defining the Pruning Objective for MHA: Method~1}

For multi-head attention (MHA) modules, we first consider an objective function that explicitly penalizes the reconstruction error of both the output projection and the intermediate attention computations. The loss is defined as:  

\begin{equation}
\begin{aligned}
& \alpha \left\| \mathbf{z}_{\ell+1}^{\text{pre}} - (\mathbf{M}^O_{\ell+1} \odot \widehat{\mathbf{W}}^O_{\ell+1}) \mathbf{a}^{attn}_{\ell} \right\|_2^2 
 \\
& + \alpha \left\| \mathbf{a}^{attn}_{\ell} - \mathrm{Concat}\bigl((\mathbf{M}^V_{\ell+1,i} \odot \widehat{\mathbf{W}}^V_{\ell+1,i}) \mathbf{a}_{\ell,i}, \ldots \bigr) \right\|_2^2 \\ 
&+ \beta \left\| \mathbf{a}_{\ell} - \phi_{\ell}(\mathbf{z}_{\ell}) \right\|_2^2 \\
&+ \alpha \left\| \mathbf{z}_{\ell} - \bigl((\mathbf{M}^Q_{\ell} \odot \widehat{\mathbf{W}}^Q_{\ell}) \mathbf{a}_{\ell-1}^{\text{pre}}\bigr) 
\bigl((\mathbf{M}^K_{\ell} \odot \widehat{\mathbf{W}}^K_{\ell}) \mathbf{a}_{\ell-1}^{\text{pre}}\bigr) \right\|_2^2,
\end{aligned}
\end{equation}

where $\mathbf{M}_\ell^Q, \mathbf{M}_\ell^K, \mathbf{M}_\ell^V, \mathbf{M}_\ell^O$ are the pruning masks and $\widehat{\mathbf{W}}_\ell^Q, \widehat{\mathbf{W}}_\ell^K, \widehat{\mathbf{W}}_\ell^V, \widehat{\mathbf{W}}_\ell^O$ denote the learnable (reconstructed) weight matrices.  

This design corresponds to the \textbf{first objective formulation} for MHA pruning. It enforces consistency between pre-trained outputs and reconstructed outputs across all projection matrices ($Q, K, V, O$), thereby encouraging pruning decisions that minimize distortion in the attention mechanism. We later compare this formulation with alternative designs to illustrate the trade-offs in optimization stability and pruning effectiveness.

We adapt the iterative pruning-update scheme used in FFN layers to the multi-head attention (MHA) modules. The procedure consists of the following steps:

\begin{itemize}
    \item \textbf{Pruning weights.}  
    Apply the same strategy as in the FFN case (Section~\ref{FFN}) to optimize the pruning masks $M_\ell$ and the corresponding reconstructed weights $\widehat{W}_\ell$.

    \item \textbf{Updating activations $\mathbf{a}_\ell$.}  
    Similar to the least-squares update in FFN, we update $\mathbf{a}_\ell$ by minimizing
    \begin{equation}
    \begin{aligned}
    \alpha \left\| \mathbf{a}^{attn}_{\ell} - \mathrm{Concat}\bigl((\mathbf{M}^V_{\ell+1,i} \odot \widehat{\mathbf{W}}^V_{\ell+1,i}) \mathbf{a}_{\ell,i}, \ldots \bigr) \right\|_2^2 + \beta \left\| \mathbf{a}_{\ell} - \phi_{\ell}(\mathbf{z}_{\ell}) \right\|_2^2.
    \end{aligned}
    \end{equation}
    Since MHA involves a non-linear $\mathrm{softmax}$ operation, this problem is not strictly linear. We therefore adopt iterative gradient-based updates (e.g., SGD solvers) to refine $\mathbf{a}_\ell$.

    \item \textbf{Updating intermediate attention activations $\mathbf{a}^{attn}_{\ell}$.}  
    Given fixed pruned weights, $\mathbf{a}^{attn}_{\ell}$ is updated by minimizing
    \begin{equation}
    \begin{aligned}
    \alpha \left\| \mathbf{z}_{\ell+1}^{\text{pre}} - (\mathbf{M}^O_{\ell+1} \odot \widehat{\mathbf{W}}^O_{\ell+1}) \mathbf{a}^{attn}_{\ell} \right\|_2^2 + 
    \alpha \left\| \mathbf{a}^{attn}_{\ell} - \mathrm{Concat}\bigl((\mathbf{M}^V_{\ell+1,i} \odot \widehat{\mathbf{W}}^V_{\ell+1,i}) \mathbf{a}_{\ell,i}, \ldots \bigr) \right\|_2^2.
    \end{aligned}
    \end{equation}

    \item \textbf{Updating outputs $\mathbf{z}_\ell$.}  
    Analogous to FFN, once weights and activations are fixed, we update $\mathbf{z}_\ell$ by minimizing
    \begin{equation}
    \begin{aligned}
    \beta \| \mathbf{a}_{\ell} - \phi_{\ell}(\mathbf{z}_{\ell}) \|_2^2 +  \alpha \Bigl\| \mathbf{z}_{\ell} - \bigl((\mathbf{M}^Q_{\ell} \odot \widehat{\mathbf{W}}^Q_{\ell}) \mathbf{a}_{\ell-1}^{\text{pre}}\bigr) 
      \bigl((\mathbf{M}^K_{\ell} \odot \widehat{\mathbf{W}}^K_{\ell}) \mathbf{a}_{\ell-1}^{\text{pre}}\bigr) \Bigr\|_2^2.
      \end{aligned}
    \end{equation}
    This update is again carried out using gradient-based solvers.

    \item \textbf{Recovering weight matrices.}  
    We estimate the weight matrices from activations via
    \begin{equation}
    \mathbf{W}_\ell = \mathbf{z}_\ell a_{\ell-1}^\dagger, 
    \quad 
    \mathbf{W}_{\ell+1} = \mathbf{z}_{\ell+1} a_\ell^\dagger.
    \end{equation}
    As computing the pseudo-inverse may be unstable or expensive, we instead use SGD-based optimization as a practical alternative.

    \item \textbf{Outer optimization loop.}  
    The optimized $\mathbf{W}_\ell$ and $\mathbf{W}_{\ell+1}$ are passed into the outer cycle. By iterating this process, we gradually obtain pruned weights that preserve model fidelity while achieving the desired sparsity.
\end{itemize}

\subsection{Defining the Pruning Objective for MHA: Method~2}

As an alternative to Method~1, we propose a second formulation of the pruning objective for multi-head attention (MHA). This version separates the constraints on $Q$, $K$, $V$, and $O$ projections, enforcing reconstruction consistency at both the attention activation and output levels. The loss function is defined as:

\begin{equation}
\begin{aligned}
& \alpha \left\| \mathbf{z}_{\ell+1}^{\text{pre}} - (\mathbf{M}^O_{\ell+1} \odot \widehat{\mathbf{W}}^O_{\ell+1}) \mathbf{a}^{attn}_{\ell} \right\|_2^2 \\
&+ \alpha \left\| \mathbf{a}^{attn}_{\ell} - \mathrm{Concat}\bigl((\mathbf{M}^V_{\ell+1,i} \odot \widehat{\mathbf{W}}^V_{\ell+1,i}) \mathbf{a}_{\ell,i}, \ldots \bigr) \right\|_2^2 \\
&+ \beta \left\| \mathbf{a}_{\ell} - \phi_{\ell}(\mathbf{z}_{\ell}) \right\|_2^2 \\
&+ \alpha \left\| \mathbf{z}_{\ell} - (\mathbf{M}^Q_{\ell} \odot \widehat{\mathbf{W}}^Q_{\ell}) \mathbf{a}_{\ell-1}^{\text{pre}} \right\|_2^2 \\
&+ \alpha \left\| \mathbf{z}_{\ell} - (\mathbf{M}^K_{\ell} \odot \widehat{\mathbf{W}}^K_{\ell}) \mathbf{a}_{\ell-1}^{\text{pre}} \right\|_2^2.
\end{aligned}
\end{equation}

Here $\mathbf{M}_\ell^Q, \mathbf{M}_\ell^K, \mathbf{M}_\ell^V, \mathbf{M}_\ell^O$ are the pruning masks and $\widehat{\mathbf{W}}_\ell^Q, \widehat{\mathbf{W}}_\ell^K, \widehat{\mathbf{W}}_\ell^V, \widehat{\mathbf{W}}_\ell^O$ are the reconstructed weights.  

The optimization procedure can be decomposed into the following steps:

\begin{itemize}[left = 0em]
    \item \textbf{Pruning weights.}  
    Optimize pruning masks and reconstructed weights using the same strategy as in the FFN case (Section~\ref{FFN}).  

    \item \textbf{Updating activations $\mathbf{a}_\ell$.}  
    Update $\mathbf{a}_\ell$ by minimizing
    \begin{equation}\label{85}
    \begin{aligned}
    \alpha \left\| \mathbf{a}^{attn}_{\ell} - \mathrm{Concat}\bigl((\mathbf{M}^V_{\ell+1,i} \odot \widehat{\mathbf{W}}^V_{\ell+1,i}) \mathbf{a}_{\ell,i}, \ldots \bigr) \right\|_2^2 
    + \beta \left\| \mathbf{a}_{\ell} - \phi_{\ell}(\mathbf{z}_{\ell}) \right\|_2^2.
    \end{aligned}
    \end{equation}
    Since MHA involves a non-linear $\mathrm{softmax}$, we apply gradient-based solvers (e.g., SGD) for iterative refinement.

    \item \textbf{Updating intermediate attention activations $\mathbf{a}^{attn}_{\ell}$.}  
    Fixing the pruned weights, we minimize
    \begin{equation}\label{86}
    \begin{aligned}
    &\alpha \left\| \mathbf{z}_{\ell+1}^{\text{pre}} - (\mathbf{M}^O_{\ell+1} \odot \widehat{\mathbf{W}}^O_{\ell+1}) \mathbf{a}^{attn}_{\ell} \right\|_2^2
    + \alpha \left\| \mathbf{a}^{attn}_{\ell} - \mathrm{Concat}\bigl((\mathbf{M}^V_{\ell+1,i} \odot \widehat{\mathbf{W}}^V_{\ell+1,i}) \mathbf{a}_{\ell,i}, \ldots \bigr) \right\|_2^2.
    \end{aligned}
    \end{equation}

    \item \textbf{Updating outputs $\mathbf{z}_\ell$.}  
    With weights and activations fixed, we minimize
    \begin{equation}\label{87}
    \begin{aligned}
    &\beta \| \mathbf{a}_{\ell} - \phi_{\ell}(\mathbf{z}_{\ell}) \|_2^2
    + \alpha \| \mathbf{z}_{\ell} - (\mathbf{M}^Q_{\ell} \odot \widehat{\mathbf{W}}^Q_{\ell}) \mathbf{a}_{\ell-1}^{\text{pre}} \|_2^2
    + \alpha \| \mathbf{z}_{\ell} - (\mathbf{M}^K_{\ell} \odot \widehat{\mathbf{W}}^K_{\ell}) \mathbf{a}_{\ell-1}^{\text{pre}} \|_2^2.
    \end{aligned}
    \end{equation}
    Gradient-based solvers are again employed for stable updates.

    \item \textbf{Recovering weight matrices.}  
    Approximate weight matrices from activations:
    \begin{equation}
    \begin{aligned}
    \mathbf{W}_\ell = \mathbf{z}_\ell a_{\ell-1}^\dagger, \quad 
    \mathbf{W}_{\ell+1} = \mathbf{z}_{\ell+1} a_\ell^\dagger.
    \end{aligned}
    \end{equation}
    Since the pseudo-inverse may be unstable, SGD-based optimization is used instead.

    \item \textbf{Outer optimization loop.}  
    The optimized weights $\mathbf{W}_\ell$ and $\mathbf{W}_{\ell+1}$ are iteratively refined through an outer optimization cycle, progressively improving pruning quality while maintaining model fidelity.
\end{itemize}

To summarize, we have extended the ADMM-based global pruning framework to both feed-forward networks (FFNs) and multi-head attention (MHA) modules. For FFN layers, the pruning objective and update rules follow a least-squares formulation, allowing efficient optimization of weights, activations, and outputs. For MHA layers, we proposed two variants of the pruning objective: Method~1 (joint consistency across $Q,K,V,O$ projections) and Method~2 (separate reconstruction constraints for each projection). In practice, we adopt \textbf{Method~2} for MHA, as it provides more stable optimization and better empirical performance.

By integrating the FFN and MHA formulations, we obtain the complete pruning framework, referred to as \textbf{SparseLLM}. The overall procedure iteratively updates pruning masks, reconstructed weights, and activations within an outer optimization loop, gradually improving pruning quality while preserving model fidelity. A detailed pseudocode of the full algorithm is provided in Algorithm~\ref{alg:sparse_llm}.

 \begin{algorithm*}[!ht]
\footnotesize
\caption{sparseLLM\_structured Pruning of OPT Models}
\label{alg:sparse_llm}
\KwIn{An OPT decoder layer containing FFN and MHA modules. Pre-trained weight matrix $W_\ell$, input of the up-scaling linear layer $a_\ell^{\text{pre}}$, output of the down-scaling linear layer $z_{\ell+1}^{\text{pre}}$, target sparsity $\rho$, constraint weight hyperparameters $\alpha, \beta$.}
\BlankLine

\SetKwFunction{SparseLLMFFN}{SparseLLM1}
\SetKwFunction{SparseLLMMHA}{SparseLLM2}
\SetKwProg{Fn}{Function}{:}{}
\Fn{\SparseLLMFFN on FFN()}{
    Initialize $z_\ell = z_\ell^{\text{pre}}, a_\ell = a_\ell^{\text{pre}}$ 
    Pre-compute and cache $a_{\ell-1}^{\text{+}} = \text{pseudo-inverse}(a_{\ell-1}^{\text{pre}})$
    \For{step $i = 1, \dots, K$}{
        $M_\ell, \hat{W}_\ell = \arg\min \|W_\ell a_{\ell-1}^{\text{pre}} - (M_\ell \odot \hat{W}_\ell)a_{\ell-1}^{\text{pre}}\|_2^2$ 
        \tcp{Prune layer $\ell$ by importance+recovery solver, see section 1}
        $M_{\ell+1}, \hat{W}_{\ell+1} = \arg\min \|W_{\ell+1}a_\ell - (M_{\ell+1} \odot \hat{W}_{\ell+1})a_\ell\|_2^2$
        \tcp{Prune layer $\ell+1$ by importance+recovery solver, see section 1}
        $W_{\ell+1} = M_{\ell+1} \odot \hat{W}_{\ell+1},  W_\ell = M_\ell \odot \hat{W}_\ell$
        $a_\ell = (\alpha W_\ell^\top W_{\ell+1} + \beta I)^{-1}(\alpha W_\ell^\top z_{\ell+1}^{\text{pre}} + \beta \phi_\beta(z_\ell))$
        $z_\ell^{(1)} = W_\ell a_{\ell-1}^{\text{pre}},  z_\ell^{(2)} = (\alpha + \beta)^{-1}(\beta a_\ell + \alpha z_\ell^{(1)})$
        \For{$j = 1, \dots, n$ \textbf{in parallel}}{
            \If{$(z_\ell)_j < 0$}{
                $(z_\ell)_j = (z_\ell^{(1)})_j$
            }
            \Else{
                $(z_\ell)_j = (z_\ell^{(2)})_j$
            }
        }
    }
    \Return{$W_\ell, W_{\ell+1}$}
}
\BlankLine
\Fn{\SparseLLMMHA on MHA()}{
    Initialize $z_\ell = z_\ell^{\text{pre}}, a_\ell = a_\ell^{\text{pre}}, a^\text{attn}_\ell = a_\ell^{\text{pre, attn}} = \mathrm{Concat}\bigl((\mathbf{M}^V_{\ell+1,i} \odot \widehat{\mathbf{W}}^V_{\ell+1,i}) \mathbf{a}^\text{pre}_{\ell,i}, \ldots \bigr)$ 
    Pre-compute and cache $a_{\ell-1}^{\text{+}} = \text{pseudo-inverse}(a_{\ell-1}^{\text{pre}})$
    \For{step $i = 1, \dots, K$}{
        $M_\ell, \hat{W}_\ell = \arg\min \|W_\ell a_{\ell-1}^{\text{pre}} - (M_\ell \odot \hat{W}_\ell)a_{\ell-1}^{\text{pre}}\|_2^2$ 
        \tcp{Prune layer $\ell$ by importance+recovery solver, do for $W_Q$, $W_K$, $W_V$, $W_O$ matrix separately. see section 1}
        $M_{\ell+1}, \hat{W}_{\ell+1} = \arg\min \|W_{\ell+1}a_\ell - (M_{\ell+1} \odot \hat{W}_{\ell+1})a_\ell\|_2^2$
        \tcp{Prune layer $\ell+1$ by importance+recovery solver, do for $W_Q$, $W_K$, $W_V$, $W_O$ matrix separately. see section 1}
        Optimize Equation \ref{85} to get: $\mathbf{a}_\ell$
        Optimize Equation \ref{86} to get: $\mathbf{a}^\text{attn}_\ell$
        Optimize Equation \ref{87} to get: $\mathbf{z}_\ell$
        \For{$j = 1, \dots, n$ \textbf{in parallel}}{
            \If{$(z_\ell)_j < 0$}{
                $(z_\ell)_j = (z_\ell^{(1)})_j$
            }
            \Else{
                $(z_\ell)_j = (z_\ell^{(2)})_j$
            }
        }
    }
    \Return{$W_\ell, W_{\ell+1}$ for $W_Q$, $W_K$, $W_V$, $W_O$ matrix separately}
}
\KwOut{Pruned model for all layers $\{W_\ell\}$.}
Initialize pruned FFN layers: $\{W_\ell^{\text{FFN}}\}$ from \SparseLLMFFN on FFN()
Initialize pruned MHA layers: $\{W_\ell^{\text{MHA}}\}$ from \SparseLLMMHA on MHA()
Combine all pruned layers into final model: $\{W_\ell^{\text{final}}\} = \{W_\ell^{\text{FFN}}\} \cup \{W_\ell^{\text{MHA}}\}$
\Return{Final pruned model $\{W_\ell^{\text{final}}\}$}

 \end{algorithm*}

\section{Solution to the Main Problem}\label{111222}

In this section, we provide the analytical solution to the main optimization objective Equation~\eqref{eq:goal} introduced earlier. Specifically, we derive closed-form or iterative update rules for the key variables under the ADMM-based structured pruning framework, which together constitute the solution to the global pruning problem.

The overall objective over \(N\) samples is
\begin{equation}
\begin{aligned}
\min_{\{r_\ell\}}
\frac{1}{N}\sum_{i=1}^N
&\sum_{\ell=1}^L\Bigl\{
\|z_\ell^{\rm pre} - (M_{\ell+1}\odot\widehat W_{\ell+1})a_\ell\|_2^2+ \|W_\ell a_{\ell-1} - (M_\ell\odot\widehat W_\ell)a_{\ell-1}^{\rm pre}\|_2^2
\Bigr\},
\end{aligned}
\end{equation}
Since we adopt structured pruning, the optimization is carried out with respect to the layer-wise sparsity ratios \(\{r_\ell\}\), rather than directly solving for element-wise pruning masks. For each layer \(\ell\), define
\begin{equation}
f(M_\ell)= \bigl\|W_\ell a_{\ell-1} -(M_\ell\odot\widehat W_\ell)a_{\ell-1}^{\rm pre}\bigr\|_2^2.
\end{equation}
Assume that \( a_{\ell-1} \), \( a_{\ell-1}^{\rm pre} \), \( W_\ell \), and \( \widehat W_\ell \) are fixed. Then both
\begin{equation}
b \coloneqq W_\ell a_{\ell-1}
\quad \text{and} \quad
c \coloneqq \widehat W_\ell a_{\ell-1}^{\rm pre}
\end{equation}
are constant vectors. As a result, the mask \( M_\ell \) operates in an element-wise manner. We write
\begin{equation}
M_\ell = \bigl(M_{\ell,1},\dots,M_{\ell,n}\bigr).
\end{equation}

we have
\begin{equation}
f(M_\ell)
= \sum_{j=1}^n \bigl(b_j - M_{\ell,j}c_j\bigr)^2.
\end{equation}
If \(M_{\ell,j}\) were relaxed to \([0,1]\), its partial derivative would be
\begin{equation}
\frac{\partial f}{\partial M_{\ell,j}}
= -2c_j\bigl(b_j - M_{\ell,j}c_j\bigr).
\end{equation}
Similarly, for
\begin{equation}
g(M_{\ell+1})
= \bigl\|z_\ell^{\rm pre} - (M_{\ell+1}\odot\widehat W_{\ell+1})a_\ell\bigr\|_2^2,
\end{equation}
we have:
\begin{equation}
g(M_{\ell})
= \sum_{j=1}^n \bigl(z^{\rm pre}_{\ell-1,j} - M_{\ell,j}d_j\bigr)^2.
\end{equation}
with 
\begin{equation}
d \coloneqq \widehat W_{\ell}a_{\ell-1},
\end{equation}
we get
\begin{equation}
\frac{\partial g}{\partial M_{\ell,j}}
= -2d_j\bigl(z_{\ell-1,j}^{\rm pre} - M_{\ell,j}d_j\bigr).
\end{equation}
Hence the total derivative of the relevant loss terms w.r.t.\ \(M_{\ell,j}\) is
\begin{equation}
\frac{\partial \mathcal{L}}{\partial M_{\ell,j}}
= -2c_j\bigl(b_j - M_{\ell,j}c_j\bigr)
  -2d_j\bigl(z_{\ell-1,j}^{\rm pre} - M_{\ell,j}d_j\bigr).
\end{equation}

subject to the sparsity constraint
\begin{equation}\label{constraint}
\sum_{\ell=1}^L \|M_\ell\|_1
= n\sum_{\ell=1}^L r_\ell = rnL.
\end{equation}
Because the loss separates over mask entries, we introduce a Lagrange multiplier \(\lambda\) for layer \(\ell\):
\begin{equation}\label{eq:78}
\begin{aligned}
\mathcal{L}(M_\ell,\lambda)
&= \sum_{j=1}^n (b_j - M_{\ell,j}c_j)^2 + \sum_{j=1}^n \bigl(z^{\rm pre}_{\ell-1,j} - M_{\ell,j}d_j\bigr)^2+ \lambda\Bigl(\sum_{j=1}^n M_{\ell,j} - r_\ell n\Bigr).
\end{aligned}
\end{equation}

For the solution of~\ref{eq:78}, Differentiate the Lagrangian with respect to $M_{\ell,j}$.

For a fixed index $ j $ the terms in the Lagrangian involving $ M_{\ell,j} $ are
\begin{equation}
f(M_{\ell,j}) = \bigl(b_j - M_{\ell,j} c_j\bigr)^2 + \bigl(z^{\rm pre}_{\ell-1,j} - M_{\ell,j} d_j\bigr)^2 + \lambda M_{\ell,j}.
\end{equation}
Taking the derivative with respect to $ M_{\ell,j} $ gives
\begin{equation}
\begin{aligned}
\frac{\partial f}{\partial M_{\ell,j}} &= -2c_j\left(b_j - M_{\ell,j}c_j\right) - 2d_j\left(z^{\rm pre}_{\ell-1,j} - M_{\ell,j}d_j\right) + \lambda \\
&= 0.
\end{aligned}
\end{equation}
Thus, solving for $ M_{\ell,j} $ we obtain
\begin{equation}\label{eq:opti}
M_{\ell,j} = \frac{c_jb_j + d_j z^{\rm pre}_{\ell-1,j} - \frac{\lambda}{2}}{c_j^2 + d_j^2}. 
\end{equation}
Substitute the Equation~\eqref{eq:opti} for $ M_{\ell,j} $:
\begin{equation}
\sum_{j=1}^n \frac{c_jb_j + d_j z^{\rm pre}_{\ell-1,j} - \frac{\lambda}{2}}{c_j^2 + d_j^2} = r_\ell n.
\end{equation}
\begin{equation}\label{lambdas}
\lambda = 2\frac{\displaystyle \sum_{j=1}^n \frac{c_jb_j + d_j z^{\rm pre}_{\ell-1,j}}{c_j^2 + d_j^2} - r_\ell n}{\displaystyle \sum_{j=1}^n \frac{1}{c_j^2 + d_j^2}}. 
\end{equation}
Substitute the Equation~\eqref{lambdas} into Equation~\eqref{eq:opti}:
\begin{equation}\label{Mlj}
\begin{aligned}
M_{\ell,j} &= \frac{c_jb_j + d_j z^{\rm pre}_{\ell-1,j} - \frac{1}{2}\times 2\displaystyle\frac{\sum_{k=1}^n \frac{c_kb_k + d_k z^{\rm pre}_{\ell-1,k}}{c_k^2 + d_k^2} - r_\ell n}{ \sum_{k=1}^n \frac{1}{c_k^2 + d_k^2}}}{c_j^2 + d_j^2}\\
 &= \frac{c_jb_j + d_j z^{\rm pre}_{\ell-1,j} - \displaystyle\frac{\sum_{k=1}^n \frac{c_kb_k + d_k z^{\rm pre}_{\ell-1,k}}{c_k^2 + d_k^2} - r_\ell n}{\sum_{k=1}^n \frac{1}{c_k^2 + d_k^2}}}{c_j^2 + d_j^2}\\
 &= \frac{1}{c_j^2+d_j^2}\left(c_jb_j + d_j z^{\rm pre}_{\ell-1,j}\right) \\
 &- \frac{1}{c_j^2+d_j^2}\cdot\frac{\sum_{k=1}^n \frac{c_kb_k + d_k z^{\rm pre}_{\ell-1,k}}{c_k^2 + d_k^2} - r_\ell n}{\sum_{k=1}^n \frac{1}{c_k^2+d_k^2}}\\
 &\quad \text{for} \quad j=1,\dots,n.
\end{aligned}
\end{equation}
the parameter $r_\ell$ appears only in the constraint term as $-\lambda r_\ell n$. Hence, the derivative with respect to $r_\ell$ is
\begin{equation}
\frac{\partial \mathcal{L}}{\partial r_\ell} = -\lambda n.
\end{equation}
Now, substituting $\lambda=0$ back into our expression for~\ref{Mlj} gives
\begin{equation}
M_{\ell,j} = \frac{c_jb_j + d_j z^{\rm pre}_{\ell-1,j}}{c_j^2+d_j^2}.
\end{equation}
we substitute in the optimal~\ref{Mlj}. Thus,
\begin{equation}
\sum_{j=1}^n \frac{c_jb_j + d_j z^{\rm pre}_{\ell-1,j}}{c_j^2+d_j^2} = r_\ell n.
\end{equation}
Solving for $r_\ell$, we obtain
\begin{equation}
r_\ell = \frac{1}{n}\sum_{j=1}^n \frac{c_jb_j + d_j z^{\rm pre}_{\ell-1,j}}{c_j^2+d_j^2}.
\end{equation}
For binary $M_{\ell,j}$ $M_{\ell,j} \in \{0,1\}$, we define a score
\begin{equation}
s_j = \frac{c_jb_j + d_j z^{\rm pre}_{\ell-1,j}}{c_j^2+d_j^2}.
\end{equation}
\begin{equation}
M_{\ell,j} = \begin{cases}
1, & \text{if } s_j > \tau,\\
0, & \text{otherwise.}
\end{cases}
\end{equation}
To construct the mask, we select the top $\tau$ scores among the $s_j$ values. This method requires jointly optimizing $M_{\ell,j}$ and $r_\ell$ to determine both the binary mask and the target sparsity. Specifically, we compute the following intermediate variables:
\begin{equation}\label{bcd}
b_j^{(\ell)} = [W_\ell a_{\ell-1}]_j,\quad
c_j^{(\ell)} = [\widehat W_\ell a_{\ell-1}^{\mathrm{pre}}]_j,\quad
d_j^{(\ell)} = [\widehat W_{\ell+1} a_\ell]_j,
\end{equation}
which are used to obtain the modified activations and auxiliary variables, capturing the adjusted dependencies that influence the final output. However, the algorithm is computationally complex due to the need for layer-wise dependency tracking and iterative mask adjustment. 

The following algorithm Algorithm~\ref{alg:GSPO} summarizes how, given fixed weights and activations, one computes the $r_\ell$ and the mask $M_{\ell,j}$, derives a Lagrange multiplier \(\lambda\), and selects the optimal entries to retain under the global sparsity budget.

 \begin{algorithm*}[!ht]
 \footnotesize
\caption{ Global Structured Pruning with Optimization}\label{alg:GSPO}
\begin{algorithmic}[1]
  \STATE \textbf{Input:} Weights $\{W_\ell\}$, pre‐activations $\{a_{\ell-1}^{\mathrm{pre}}\}$, activations $\{a_{\ell-1}\}$, sparsity $r$
\FOR{$\text{step } i = 1, \dots, K$}
    \STATE $W_\ell = \mathbf{z}_\ell a_{\ell-1}^\dagger$, \quad $W_{\ell+1} = \mathbf{z}_{\ell+1} a_\ell^\dagger$
\FOR{$\ell=1$ \TO $L$}
  \STATE For each $\ell=1,\dots,L$ and $j=1,\dots,n$, compute
    \begin{equation}
      b_j^{(\ell)} = [W_\ell a_{\ell-1}]_j,\quad
      c_j^{(\ell)} = [\widehat W_\ell a_{\ell-1}^{\mathrm{pre}}]_j,\quad
      d_j^{(\ell)} = [\widehat W_{\ell+1} a_\ell]_j.
    \end{equation}
  \STATE Compute $M_{\ell,j}$ and $r_\ell$
    \begin{equation}
      M_{\ell,j}= \frac{1}{c_j^2+d_j^2}\left(c_jb_j + d_j z^{\rm pre}_{\ell-1,j}\right) - \frac{1}{c_j^2+d_j^2}\cdot\frac{\sum_{k=1}^n \frac{c_kb_k + d_k z^{\rm pre}_{\ell-1,k}}{c_k^2 + d_k^2} - r_\ell n}{\sum_{k=1}^n \frac{1}{c_k^2+d_k^2}},
    \end{equation}
 \begin{equation}
    r_\ell = \frac{1}{n}\sum_{j=1}^n \frac{c_jb_j + d_j z^{\rm pre}_{\ell-1,j}}{c_j^2+d_j^2}.
 \end{equation} 
  \STATE Collect all $\{s_{\ell,j}\}$, sort in descending order, let $K=\lfloor r n L\rfloor$, and set threshold $\tau$ to the $K$th largest score.
\FOR{$j = 1$ \TO $n$} 
    \STATE $M_{\ell,j} \gets \mathbb{I}\{s_{\ell,j} > \tau\}$
\ENDFOR
    \STATE $W_{\ell+1} = M_{\ell+1} \odot \widehat{W}_{\ell+1}$,\quad $W_\ell = M_\ell \odot \widehat{W}_\ell$
    \STATE $a_\ell = \left(\alpha W_{\ell+1}^\top W_{\ell+1} + \beta I \right)^{-1} \left( \alpha W_{\ell+1}^\top \mathbf{z}_{\ell+1}^{\text{pre}} + \beta \phi_\ell(\mathbf{z}_\ell) \right)$ 
    \STATE $\mathbf{z}_\ell^{(1)} = W_\ell a_{\ell-1}^{\text{pre}}$
    \STATE $\mathbf{z}_\ell^{(2)} = (\alpha + \beta)^{-1} \cdot \left( \beta a_\ell + \alpha \mathbf{z}_\ell^{(1)} \right)$
\STATE \begin{equation}
(\mathbf{z}_\ell)_j =
\begin{cases}
(\mathbf{z}_\ell^{(1)})_j & \text{if } (\mathbf{z}_\ell)_j < 0, \\
(\mathbf{z}_\ell^{(2)})_j & \text{otherwise},
\end{cases}
\quad \text{for } j = 1, \dots, n \text{ (in parallel)}.
\end{equation}
\ENDFOR
\ENDFOR
  \STATE \textbf{Output:} Masks $\{M_\ell\}$ satisfying $\sum_{\ell=1}^L\sum_{j=1}^n M_{\ell,j} = r n L$.
\end{algorithmic}
 \end{algorithm*}

\section{Energy-based Asymptotic Framework}\label{energy}

Given precomputed constants \( b_j, c_j, d_j \) from Equation~\eqref{bcd}, the original loss (Equation~\eqref{eq:goal} at layer \( \ell \) is defined as:
\begin{equation}
\begin{aligned}
\mathcal{L}(M_\ell,\lambda) = \sum_{j=1}^n (b_j - M_{\ell,j}c_j)^2 + \sum_{j=1}^n \left(z^{\mathrm{pre}}_{\ell-1,j} - M_{\ell,j}d_j\right)^2  +\lambda \left(\sum_{j=1}^n M_{\ell,j} - r_\ell n\right).
\end{aligned}
\end{equation}
In many pruning or sparsification schemes one is not interested in the detailed behavior of every $M_{\ell,j}$ but rather in the overall fraction of active units $r_\ell$ for the layer. In other words, rather than optimizing over the detailed variables, one wishes compute an effective cost per layer as a function of the sparsity level. 

By assuming  mean-field/large–deviation arguments~\cite{mei2018mean} that there is an averaging effect when the number of neurons $n$ is large, the loss per layer can be summarized by an energy function $E_\ell(r_\ell)$. In many statistical–mechanics inspired approaches~\cite{sirignano2020mean}, one finds a logarithmic dependence when relating the number of accessible configurations (here, possible choices of neurons to keep) with the fraction of neurons kept. In fact, counting the available state gives a combinatorial factor that has the asymptotic form.

Rather than optimizing over binary variables \( M_{\ell,j} \), we approximate their collective behavior using a large-deviation or mean-field argument~\cite{mei2018mean,sirignano2020mean}. See Section~\ref{stirling} for details. It leads to an effective energy function \( E_\ell(r_\ell) \) that depends only on the average retention rate \( r_\ell \). The total relaxed objective becomes:
\begin{equation}
\mathcal{L}_{\text{relaxed}} = \sum_{\ell=1}^L E_\ell(r_\ell) + \lambda \left( \sum_{\ell=1}^L r_\ell - rL \right).
\end{equation}
Assume that exactly \( r_\ell n \) out of the \( n \) units in layer \( \ell \) are retained (i.e., have \( M_{\ell,j} = 1 \)). The number of binary configurations \( N(r_\ell) \) consistent with this constraint is given by the binomial coefficient:
\begin{equation}
N(r_\ell) = \binom{n}{r_\ell n}.
\end{equation}
Applying Stirling’s formula~\cite{stirling1730}, for large \( n \) we have:
\begin{equation}
\binom{n}{r_\ell n} \sim \exp\left\{ n H(r_\ell) \right\},
\end{equation}
where the per-neuron entropy is:
\begin{equation}
H(r_\ell) = -r_\ell \log r_\ell - (1-r_\ell)\log(1 - r_\ell).
\end{equation}
For regimes where \( r_\ell \ll 1 \), the second term can be expanded as:
\begin{equation}
(1 - r_\ell)\log(1 - r_\ell) = -r_\ell + \mathcal{O}(r_\ell^2),
\end{equation}
yielding the approximation:
\begin{equation}
H(r_\ell) = -r_\ell \log r_\ell + r_\ell.
\end{equation}
Neglecting the additive linear term (which can be absorbed into a scale), the dominant behavior is:
\begin{equation}
\log N(r_\ell) \sim n \cdot \left( -r_\ell \log r_\ell \right).
\end{equation}
To define a per-layer energy function from this combinatorial structure, we associate the number of accessible configurations with an entropy-induced energy by setting:
\begin{equation}
E_\ell(r_\ell) := -\log N(r_\ell) \sim \beta_\ell \log \frac{1}{r_\ell} = -\beta_\ell \log r_\ell,
\end{equation}
where \( \beta_\ell > 0 \) serves as an inverse temperature or penalty strength. Inspired by information-theoretic interpretations, we define:
\begin{equation}
\beta_\ell = e^{-I_\ell / T},
\end{equation}
where \( I_\ell \) is a layer-wise importance score and \( T \) is a temperature hyperparameter that controls the sensitivity of the pruning cost to the retention rate.

This results in the total relaxed pruning cost:
\begin{equation}
\mathcal{L}_{\text{relaxed}} = \sum_{\ell=1}^L -\beta_\ell \log r_\ell + \lambda \left( \sum_{\ell=1}^L r_\ell - rL \right),
\end{equation}
which enables efficient optimization over continuous variables \( r_\ell \in (0,1] \) rather than discrete binary masks.

Let $E_\ell(r_\ell) = -\beta_\ell \log r_\ell$ with $\beta_\ell = e^{-I_\ell/T}$, where $I_\ell$ denotes the layer importance and $T > 0$ is a temperature parameter. Then
\begin{equation}
    E_\ell'(r_\ell) = -\frac{\beta_\ell}{r_\ell} = -\frac{e^{-I_\ell/T}}{r_\ell}.
\end{equation}
The optimality condition yields
\begin{equation}
    r_\ell^* = \frac{e^{-I_\ell/T}}{\lambda}.
\end{equation}
Using the constraint $\sum_{\ell=1}^L r_\ell = rL$, we obtain
\begin{equation}
    \sum_{\ell=1}^L \frac{e^{-I_\ell/T}}{\lambda} = rL \quad \Longrightarrow \quad \lambda = \frac{1}{rL} \sum_{\ell=1}^L e^{-I_\ell/T}.
\end{equation}
Therefore, the optimal retention rate becomes
\begin{equation}
    r_\ell^* = rL \cdot \frac{e^{-I_\ell/T}}{\sum_{j=1}^L e^{-I_j/T}}.
\end{equation}
This is equivalent to
\begin{equation}
    r_\ell^* = rL \cdot \mathrm{softmax}(-I_\ell/T),
\end{equation}
which naturally induces a softmax allocation over the layer importance scores.

The formulation assigns higher retention to layers with lower $I_\ell$, modulated by the temperature $T$. A smaller $T$ results in more concentrated allocations. The energy function $E_\ell(r_\ell) = -e^{-I_\ell/T} \log r_\ell$ thus provides a principled route to softmax-based allocation via constrained optimization.

\section{Stirling's Approximation and Its Applications in Mean-Field Theory}\label{stirling}

\subsection{Stirling's Formula}

For large integers \( n \), Stirling's approximation provides an asymptotic expression for the factorial:
\begin{equation}
n! = \sqrt{2\pi n} \left( \frac{n}{e} \right)^n.
\end{equation}
In statistical physics, the logarithmic form is more commonly used:
\begin{equation}
\log n! = n \log n - n + \frac{1}{2} \log (2\pi n).
\end{equation}
In mean-field approximations, constant terms are often neglected, and only the leading-order terms are retained:
\begin{equation}
\log n! = n \log n - n.
\end{equation}
\subsection{Application to Entropy in Multiparticle Systems}

Consider a discrete system where particles are distributed among several distinct states. Let \( n_i \) denote the number of particles in state \( i \), and let the total number of particles be \( N = \sum_i n_i \). The total number of microstates corresponding to such a configuration is given by the multinomial coefficient:
\begin{equation}
\Omega = \frac{N!}{\prod_i n_i!}.
\end{equation}
To compute the entropy, we evaluate
\begin{equation}
S = k_B \log \Omega.
\end{equation}
Using Stirling's approximation~\cite{stirling1730} and omitting constant terms, we obtain:
\begin{equation}\begin{aligned}
\log \Omega &= N \log N - N - \sum_i \left( n_i \log n_i - n_i \right) \\
&= - \sum_i n_i \log \left( \frac{n_i}{N} \right),
\end{aligned}\end{equation}
which leads to an entropy expression of the form
\begin{equation}
S = -k_B \sum_i n_i \log \left( \frac{n_i}{N} \right).
\end{equation}
This is the discrete form of the Shannon entropy and is widely used in mean-field theory, maximum entropy models, and information-theoretic formulations of statistical mechanics.

\subsection{Mean-Field Free Energy and Partition Function}

In statistical mean-field theory, such as in the uniform mean-field approximation of the Ising model, we often define a free energy functional:
\begin{equation}
F[m] = U[m] - T S[m],
\end{equation}
where \( m \in [-1,1] \) denotes the magnetization, \( U[m] \) is the internal energy, and \( S[m] \) is the entropy term derived using Stirling’s approximation.

Let \( N_+ \) and \( N_- = N - N_+ \) denote the number of up and down spins, respectively. The number of microstates is given by the binomial coefficient:
\begin{equation}
\log \Omega = \log \binom{N}{N_+} = - N_+ \log \left( \frac{N_+}{N} \right) - N_- \log \left( \frac{N_-}{N} \right).
\end{equation}
Using the magnetization definition \( m = \frac{N_+ - N_-}{N} \), we can express the entropy as a function of \( m \), and the resulting mean-field free energy becomes:
\begin{equation}
\begin{aligned}
F(m) &= -J m^2 + T \Bigg[ \frac{1+m}{2} \log \left( \frac{1+m}{2} \right) + \frac{1-m}{2} \log \left( \frac{1-m}{2} \right) \Bigg],
\end{aligned}
\end{equation}
where \( J \) is the interaction strength. This form is derived directly from the application of Stirling's approximation and provides a tractable expression for analyzing phase behavior under the mean-field assumption.

\end{document}